\documentclass[conference]{IEEEtran}
\IEEEoverridecommandlockouts
\usepackage{cite}
\usepackage{amsmath,amssymb,amsfonts}
\usepackage{algorithmic}
\usepackage{graphicx}
\usepackage{textcomp}
\usepackage{xcolor}
\usepackage{booktabs}
\usepackage{makecell} 
\usepackage{multirow}  
\usepackage{colortbl}
\usepackage{mathtools}

\def\BibTeX{{\rm B\kern-.05em{\sc i\kern-.025em b}\kern-.08em
    T\kern-.1667em\lower.7ex\hbox{E}\kern-.125emX}}
\begin{document}

\title{Decoupled Sensitivity-Consistency Learning for Weakly Supervised Video Anomaly Detection}

\author{
    \IEEEauthorblockN{Hantao Zheng$^{1}$, Ning Han$^{2*}$, Yawen Zeng$^{1}$, Hao Chen$^{1}$}
    \IEEEauthorblockA{$^{1}$School of Computer Science, Hunan University, China}
    \IEEEauthorblockA{$^{2}$School of Computer Science, Xiangtan University, China \vspace{-1.0cm} }
    \thanks{$^{*}$Corresponding author.}
}

\maketitle

\begin{abstract}
Recent weakly supervised video anomaly detection methods have achieved significant advances by employing unified frameworks for joint optimization. 
%
However, this paradigm is limited by a fundamental sensitivity–stability trade-off, as the conflicting objectives for detecting transient and sustained anomalies lead to either fragmented predictions or over-smoothed responses.
To address this limitation, we propose DeSC, a novel \textbf{De}coupled \textbf{S}ensitivity-\textbf{C}onsistency framework that trains two specialized streams using distinct optimization strategies. 
The temporal sensitivity stream adopts an aggressive optimization strategy to capture high-frequency abrupt changes, whereas the semantic consistency stream applies robust constraints to maintain long-term coherence and reduce noise. 
Their complementary strengths are fused through a collaborative inference mechanism that reduces individual biases and produces balanced predictions. 
Extensive experiments demonstrate that DeSC establishes new state-of-the-art performance by achieving 89.37\%AUC on UCF-Crime (+1.29\%) and 87.18\%AP on XD-Violence (+2.22\%).
Code is available at https://github.com/imzht/DeSC.

\end{abstract}


\section{Introduction}
\label{sec_intro}

Video anomaly detection (VAD) has been widely studied recently, driven by its importance in real-world applications such as intelligent surveillance~\cite{sultani2018real} and video content moderation\cite{wu2020not}. While fully supervised approaches have yielded strong performance, their reliance on labor-intensive frame-level annotations limits their scalability in large-scale deployments. Consequently, weakly supervised video anomaly detection (WSVAD) has emerged as a practical alternative, as it requires only coarse video-level labels during training. 
Nevertheless, detecting frame-level anomalies in complex scenes remains challenging without precise temporal guidance.

\begin{figure}[t]
\centering
\includegraphics[width=1\columnwidth]{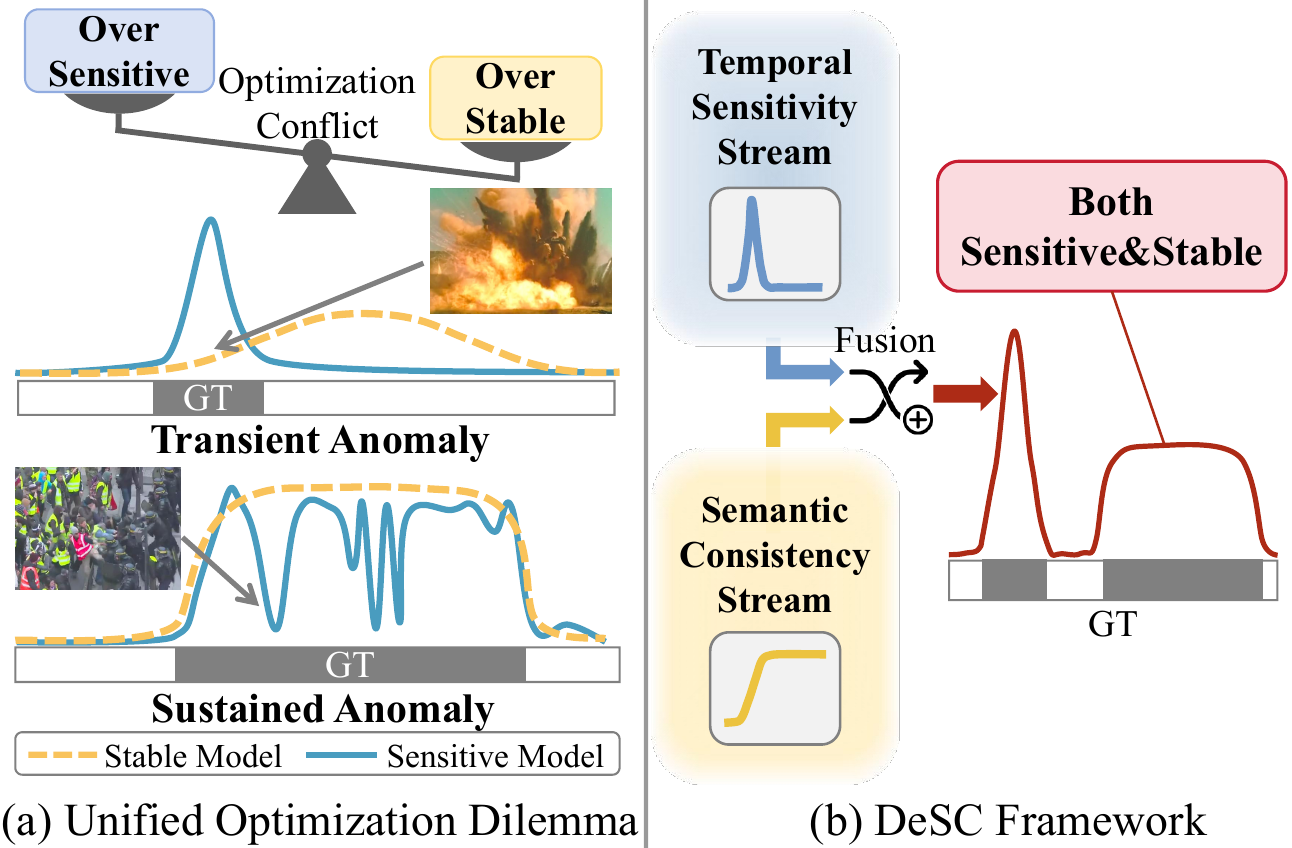}
\caption{
The sensitivity–stability trade-off. 
(a) Unified Optimization Dilemma. Prioritizing sensitivity yields fragmented predictions (bottom), whereas enforcing stability leads to over-smoothed responses (top). 
(b) DeSC resolves this via two decoupled and collaborative streams. 
GT denotes ground truth. 
}
\label{fig:teaser}
\end{figure}

Weakly supervised methods have evolved from simple classification toward advanced  feature learning paradigms. Early efforts RTFM\cite{tian2021weakly} assume that anomalous features have larger magnitudes than normal ones, thus combining a magnitude-based learning objective with a top-k multiple instance learning strategy. To capture more structured prototype patterns, DMU\cite{zhou2023dual} introduces dual memory units that explicitly maintain normal and abnormal prototypes while leveraging global-local attention to model their dependencies. More recent approaches, including VadCLIP\cite{wu2024vadclip} and STPrompt\cite{wu2024weakly}, incorporate semantic priors from vision-language models to guide the detection process. Despite these diverse designs, a common limitation remains in these state-of-the-art approaches. Most methods adopt a unified end-to-end framework and rely on a joint optimization strategy with a single set of hyperparameters to learn all representations. This common approach implicitly assumes that a single model is sufficient to handle the diverse and complex nature of real-world anomalies.

However, this unified strategy is limited by a sensitivity–stability trade-off, as optimizing for transient anomalies produces fragmented predictions while optimizing for sustained events yields over-smoothed responses, as shown in Fig.~\ref{fig:teaser}(a). 
Real-world anomalies exhibit diverse temporal characteristics where some events are abrupt and short while others are continuous and evolve over long periods. Capturing abrupt anomalies requires high temporal sensitivity to detect rapid frame-wise changes. Conversely, detecting continuous anomalies necessitates strong contextual stability to maintain prediction consistency and avoid fragmentation. Our experiments reveal that attempting to optimize these conflicting objectives within a single unified framework leads to significant gradient conflicts. A sensitive model optimized for abrupt changes is oversensitive to noise and produces fragmented predictions. In contrast, a stable model optimized for smoothness suffers from a delayed response and over-smooths short-term anomalies. This dilemma traps unified models in a sub-optimal balance where neither sensitivity nor stability is maximized.

To overcome this challenge, we propose \textbf{DeSC}, a novel \textbf{De}coupled \textbf{S}ensitivity-\textbf{C}onsistency framework. As shown in Fig.~\ref{fig:teaser}(b), DeSC shifts the paradigm from joint optimization to decoupled optimization and collaborative inference. Instead of seeking a compromise within a single model, DeSC constructs two specialized streams trained independently with task-specific objectives. The temporal sensitivity stream uses a parallel framework combining temporal convolutional networks and graph transformers to capture high-frequency abrupt anomalies. Meanwhile, the semantic consistency stream employs a graph convolutional network enhanced with Gaussian mixture priors to ensure the temporal continuity and semantic coherence of long-duration events. During inference, DeSC fuses these complementary streams through a collaborative mechanism. This strategy effectively mitigates their weaknesses by suppressing the noise from the sensitive stream while compensating for the insensitivity of the consistent stream.

Our main contributions are summarized as follows:
\begin{itemize}
\item We identify the sensitivity–stability trade-off in WSVAD, revealing that the optimization conflict between transient and sustained anomaly detection is a primary bottleneck for existing unified frameworks.
\item We propose DeSC, a decoupled framework that independently optimizes temporal sensitivity and semantic consistency. This paradigm allows each stream to achieve its best performance without mutual conflict.
\item DeSC achieves new state-of-the-art performance on UCF-Crime and XD-Violence. 
Notably, even a single decoupled stream surpasses prior state-of-the-art methods, demonstrating that the decoupled optimization strategy effectively unlocks performance potential.
\end{itemize}

\begin{figure*}[ht]
\centering

\includegraphics[width=0.95\textwidth]{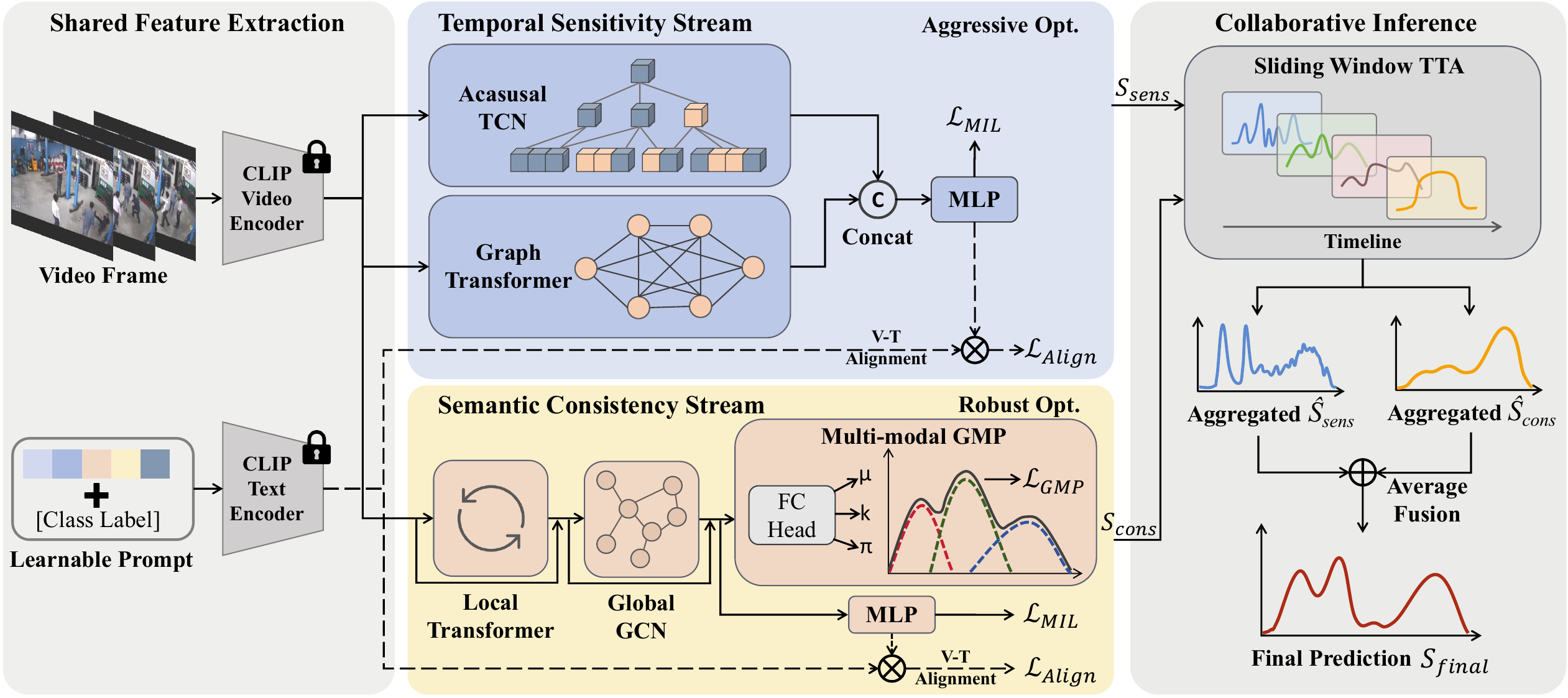} 
\caption{Overview of the DeSC framework. DeSC uses frozen CLIP features and two decoupled streams optimized with distinct objectives. The Temporal Sensitivity Stream applies an acausal TCN and graph transformer to capture transient anomalies under aggressive optimization, while the Semantic Consistency Stream uses a local transformer, global GCN, and multimodal Gaussian prior to maintain stable long-range semantics under robust optimization. Sliding-window test-time augmentation and collaborative fusion integrate both outputs for final anomaly scores.}
\label{fig:framework}
\end{figure*}
\section{Related work}
Weakly supervised video anomaly detection has evolved from visual-only multiple instance learning\cite{sultani2018real} approaches to vision-language paradigms. Early methods such as RTFM\cite{tian2021weakly} and UR-DMU\cite{zhou2023dual} improve feature discriminability through contrastive objectives or memory mechanisms but lack semantic guidance. 
Vision–language models have enabled effective cross-modal alignment for VAD, as seen in VadCLIP\cite{wu2024vadclip} and STPrompt\cite{wu2024weakly}, which use learnable prompts to connect visual features with textual anomaly descriptions.
Recent works further expand this area with Anomize\cite{li2025anomize} addressing open-vocabulary ambiguity via group-guided text encoding, while VERA\cite{ye2025vera} and Holmes-VAD\cite{zhang2024holmes} leverage verbalized learning and multi-modal LLMs for explainable anomaly detection.
Although these architectures advance cross-modal reasoning, most rely on a unified optimization pipeline that forces a compromise between the needs of transient and sustained anomalies. We address this limitation with DeSC, which separates these objectives into two independent streams and achieves collaborative inference without optimization conflict.
\section{Method}

\subsection{Overview}
Weakly supervised video anomaly detection aims to predict frame-level anomaly scores using only video-level binary labels and inferring precise boundaries from coarse-grained supervision.
To address the inherent sensitivity-stability trade-off, we propose DeSC, a framework that shifts from joint optimization to a decoupled strategy. As illustrated in Fig.~\ref{fig:framework}, we train two specialized streams under a decoupled optimization framework comprising a temporal sensitivity stream to capture high-frequency transient anomalies and a semantic consistency stream focused on enforcing the completeness of sustained events. Finally, a collaborative inference strategy combines their complementary strengths to achieve robust detection.

\subsection{Temporal Sensitivity Stream}
Capturing transient anomalies such as explosions requires a model highly sensitive to abrupt temporal variations. Conventional sequential frameworks often weaken these high-frequency signals due to layer-wise feature smoothing. 
To maximize sensitivity, we design a parallel architecture that captures local temporal dynamics and global dependencies independently. 
Given the visual features $X \in \mathbb{R}^{T \times D}$ extracted by a frozen CLIP encoder, we add learnable positional embeddings to preserve temporal order. The stream then splits into two parallel branches to extract complementary features. 
We formulate this parallel feature extraction process as: 
\begin{equation}
X_{local} = \text{TCN}(X), \quad X_{global} = \text{GT}(X, A),
\end{equation}
where $\text{TCN}(\cdot)$ comprises stacked residual dilated convolution blocks that rapidly expand the receptive field while preserving temporal resolution, and $\text{GT}(\cdot)$ represents a graph transformer that models long-range global dependencies by injecting a distance-based adjacency matrix $A$ into the self-attention mechanism. 
To encode temporal distance, the elements of $A$ are defined by an exponential decay function, formulated as

\begin{equation}
A_{i,j} = \exp\left(-\frac{|i-j|}{\tau}\right),
\end{equation}
where $|i-j|$ denotes the temporal distance between snippet $i$ and $j$, and $\tau$ is a scale parameter that controls the sparsity of the graph.
Unlike serial connections that may blur high-frequency details, our parallel design preserves local feature sensitivity by fusing the concatenated outputs via a non-linear projection, which is given by

\begin{equation}
X_{sens} = \text{MLP}(\text{Concat}(X_{local}, X_{global})),
\end{equation}
yielding the sensitivity-oriented feature representation $X_{sens}$, which is then fed into a classifier to generate the frame-level anomaly score $S_{sens}$. 
We optimize a joint objective consisting of the standard MIL classification loss $\mathcal{L}_{MIL}$ and a visual-text alignment loss $\mathcal{L}_{Align}$, ensuring both accurate detection and semantic consistency with anomaly prompts.

\subsection{Semantic Consistency Stream}
Sustained anomalies, such as burglary, suffer from fragmentation due to the inherent noise in weakly supervised settings, requiring a strong structural prior to preserve event completeness. To address this, we design a semantic consistency stream that employs a hybrid backbone that combines a local transformer and a global graph convolutional network to capture comprehensive semantic relationships among video snippets. To enforce temporal coherence, we incorporate a multi-modal Gaussian Mixture Prior (GMP) mechanism that models the temporal distribution of anomalies as a continuous mixture of Gaussian components. Specifically, the network first predicts frame-level parameters, which are then aggregated via temporal average pooling to obtain video-level Gaussian parameters, the center $\mu_k$ and temporal scale $\sigma_k$ for $K$ potential anomaly events. These parameters construct the prior distribution, which is given by

\begin{equation}
\mathcal{G}(t) = \sum_{k=1}^{K} \pi_k(t) \cdot \exp\left(-\frac{(t - \mu_k)^2}{2\sigma_k^2}\right),
\end{equation}
where $t$ represents the temporal index. The weight $\pi_k(t)$ is dynamically derived from the visual-text alignment scores, ensuring that the prior distribution is weighted by the semantic confidence at each moment. To align the model predictions with this smooth structural prior, we apply a consistency loss $\mathcal{L}_{GMP}$ that minimizes the deviation between the classification score $S_{cons}(t)$ and the distribution $\mathcal{G}(t)$, formulated as
\begin{equation}
\mathcal{L}_{GMP} = \| S_{cons} - \mathcal{G} \|_2^2.
\end{equation}



Consequently, total objective for this stream combines classification, alignment, and regularization losses, calculated as
\begin{equation}
\mathcal{L}_{total} = \mathcal{L}_{MIL} + \mathcal{L}_{Align} + \lambda \mathcal{L}_{GMP}.
\end{equation}

During training, we minimize this total objective to jointly optimize semantic consistency and detection accuracy.

\subsection{Decoupled Optimization Paradigm}
Optimizing sensitivity and consistency in a unified framework leads to gradient conflicts as the temporal sensitivity stream requires rapid parameter updates to fit abrupt changes, whereas the semantic consistency stream demands stable updates to maintain distribution shapes. To resolve this dilemma, we propose a decoupled training paradigm where each stream is optimized in a distinct environment designed for its specific dynamic requirements. For the temporal sensitivity stream, we employ an aggressive optimization strategy using a high learning rate which allows the model to escape local minima and accurately fit transient signals. Conversely for the semantic consistency stream we adopt a robust optimization strategy with a lower learning rate to ensure stable convergence towards the Gaussian prior while preserving semantic patterns. This isolation ensures that both streams achieve optimal performance without the interference of conflicting gradients.

\subsection{Collaborative Inference}
While decoupled training allows each expert stream to reach its performance limit, relying on a single stream during inference exhibits its specific bias, resulting in either fragmented or over-smoothed predictions. To combine their complementary strengths, we introduce a collaborative inference mechanism using test-time augmentation (TTA). First, to mitigate the boundary effects caused by fixed-length video cropping, we employ a sliding window strategy with temporal overlap instead of non-overlapping division. Let $\mathcal{W}$ be the set of all sliding windows covering the video. The robust score for a time step $n$ in a stream is aggregated by normalizing the predictions from all windows that encompass $n$, computed as

\begin{equation}
    \hat{S}_{stream}(n) = \frac{\sum\limits_{w \in \mathcal{W}} S_{stream}^{(w)}(n) \cdot \mathbb{I}(n \in w)}{\sum\limits_{w \in \mathcal{W}} \mathbb{I}(n \in w)},
\end{equation}
where $\mathbb{I}(\cdot)$ is the indicator function which equals 1 if time step $n$ is within window $w$ and 0 otherwise, $S_{stream}^{(w)}(n)$ denotes the prediction from window $w$, and stream $\in \{sens, cons\}$. This aggregation process ensures smooth transitions across temporal boundaries and provides a stable estimation for long-duration videos. Finally, the global anomaly score is derived by the collaborative fusion of expert streams, given by
\begin{equation}
S_{final}(n) = \frac{1}{2} \left( \hat{S}_{sens}(n) + \hat{S}_{cons}(n) \right).
\end{equation}

By fusing the complementary response patterns, DeSC reduces the fragmentation noise from the sensitivity stream and the potential insensitivity of the consistency stream, achieving robust detection that outperforms either stream individually.
\begin{table}[t]
\centering
\small
\caption{
\textbf{Comparison with state-of-the-art methods on UCF-Crime and XD-Violence datasets.} 
Best results are in \textbf{bold}.
}
\label{tab:sota_comparison}
\setlength{\tabcolsep}{8pt}
\begin{tabular}{l c c c}
\toprule
\textbf{Method} & \textbf{Feature} & \textbf{\makecell{UCF \\ (AUC \%)}} & \textbf{\makecell{XD \\ (AP \%)}} \\
\midrule

Sultani~et~al.\cite{sultani2018real} & I3D & 75.41 & 73.20 \\
RTFM\cite{tian2021weakly} & I3D & 84.30 & 77.81 \\
LA-Net\cite{pu2022locality} & I3D & 85.12 & 80.72 \\
MSL\cite{Li_Liu_Jiao_2022} & I3D & 85.30 & 78.28 \\
MGFN\cite{chen2023mgfn} & I3D & 86.98 & 80.11 \\
UR-DMU\cite{zhou2023dual} & I3D & 86.97 & 81.66 \\

\midrule
Sultani~et~al.\cite{sultani2018real} & CLIP & 84.14 & 75.18 \\
AVVD\cite{wu2022weakly} & CLIP & 82.45 & 78.10 \\
RTFM\cite{tian2021weakly} & CLIP & 85.66 & 78.27 \\
UMIL\cite{lv2023unbiased} & CLIP & 86.75 & - \\
UR-DMU\cite{zhou2023dual} & CLIP & 86.75 & 82.41 \\
CLIP-TSA\cite{joo2023clip} & CLIP & 87.58 & 82.17 \\
TPWNG\cite{yang2024text} & CLIP & 87.79 & 83.68 \\
VadCLIP\cite{wu2024vadclip} & CLIP & 88.02 & 84.51 \\
STPrompt\cite{wu2024weakly} & CLIP & \underline{88.08} & 83.97 \\
Holmes-VAD\cite{zhang2024holmes} & CLIP & 84.61 & \underline{84.96} \\
VERA\cite{ye2025vera} & CLIP & 86.55 & 70.54 \\

\midrule
\rowcolor{gray!30}
\textbf{DeSC (Ours)} & \textbf{CLIP} & \textbf{89.37} & \textbf{87.18} \\
\bottomrule
\end{tabular}
\end{table}

\section{Experiments}

\subsection{Datasets and Evaluation Metrics}
We evaluate DeSC on two large-scale video anomaly detection benchmarks. 
\textbf{UCF-Crime}~\cite{sultani2018real} contains 1,900 untrimmed surveillance videos across 13 real-world anomaly categories. 
\textbf{XD-Violence}~\cite{wu2020not} comprises 4,754 videos from movies and online platforms, covering 6 categories of violent events under diverse conditions. 
Following standard protocols~\cite{wu2024vadclip,lv2023unbiased}, we report the Area Under the Curve (AUC) on UCF-Crime and Average Precision (AP) on XD-Violence.

\begin{table}[h]
\centering
\small
\renewcommand{\arraystretch}{1}
\setlength{\tabcolsep}{8pt}
\caption{
\textbf{Ablation study on the optimization paradigm.}
We compare the joint training strategy against our decoupled optimization strategy.
``Unified'' denotes training the TCN, GT, and GMP modules jointly.
``Temp.'' and ``Sem.'' denote the Temporal Sensitivity and Semantic Consistency streams.
}
\label{tab:ablation_optimization}
\begin{tabular}{l c c c}
\toprule
\textbf{Method} & \textbf{Opt. Strategy} & \textbf{\makecell{UCF \\ (AUC \%)}} & \textbf{\makecell{XD \\ (AP \%)}} \\
\midrule

Unified & Joint & 86.18 & 80.22 \\

\midrule

\textbf{Temp. Stream} & Decoupled  & 88.46 & 85.04 \\
\textbf{Sem. Stream} & Decoupled  & 88.35 & 85.48 \\
\rowcolor{gray!30}
\textbf{DeSC} & \textbf{Collaborative} & \textbf{89.37} & \textbf{87.18} \\
\bottomrule
\end{tabular}
\end{table}

\begin{table}[h]
\centering
\small
\renewcommand{\arraystretch}{0.9}
\caption{
\textbf{Effectiveness of the collaborative inference strategy.} 
We compare the basic ensemble with the proposed test-time augmentation (TTA) using sliding windows.
}
\label{tab:ablation_inference}
\setlength{\tabcolsep}{5pt}
\begin{tabular}{l c c c}
\toprule
\textbf{Method} & \textbf{Inference Strategy} & \textbf{\makecell{UCF \\ (AUC \%)}} & \textbf{\makecell{XD \\ (AP \%)}} \\
\midrule

DeSC & Basic Ensemble & 89.13 & 86.47 \\

\rowcolor{gray!30}
\textbf{DeSC} & \textbf{w/ TTA (Sliding Window)} & \textbf{89.37} & \textbf{87.18} \\
\bottomrule
\end{tabular}
\end{table}

\begin{table}[h]
\centering
\small
\caption{
\textbf{Ablation study on internal module designs.} 
We evaluate the contribution of key components within each stream using their respective optimal training protocols.
}
\label{tab:ablation_internal}
\setlength{\tabcolsep}{5pt}
\begin{tabular}{l c c}
\toprule
\textbf{Module Configuration} & \textbf{\makecell{UCF \\ (AUC \%)}} & \textbf{\makecell{XD \\ (AP \%)}} \\
\midrule
\multicolumn{3}{l}{\textit{\underline{Temporal Sensitivity Stream}}} \\[6pt]
TCN Only & 85.80 & 79.78 \\
Graph Transformer (GT) Only & 86.26 & 84.55 \\
\textbf{Parallel TCN + GT (Ours)} & \textbf{88.46} & \textbf{85.04} \\

\midrule
\multicolumn{3}{l}{\textit{\underline{Semantic Consistency Stream}}} \\[6pt]
GCN Baseline (w/o GMP) & 87.70 & 84.67 \\
\textbf{GCN + GMP (Ours)} & \textbf{88.35} & \textbf{85.48} \\
\bottomrule
\end{tabular}
\end{table}

\definecolor{c_gt}{HTML}{383838}      %
\definecolor{c_temporal}{HTML}{3D8AA9} %
\definecolor{c_ours}{HTML}{B52B23}     %
\definecolor{c_semantic}{HTML}{D49A3A} %

\begin{figure*}[t]
  \centering
  \includegraphics[width=0.95\linewidth]{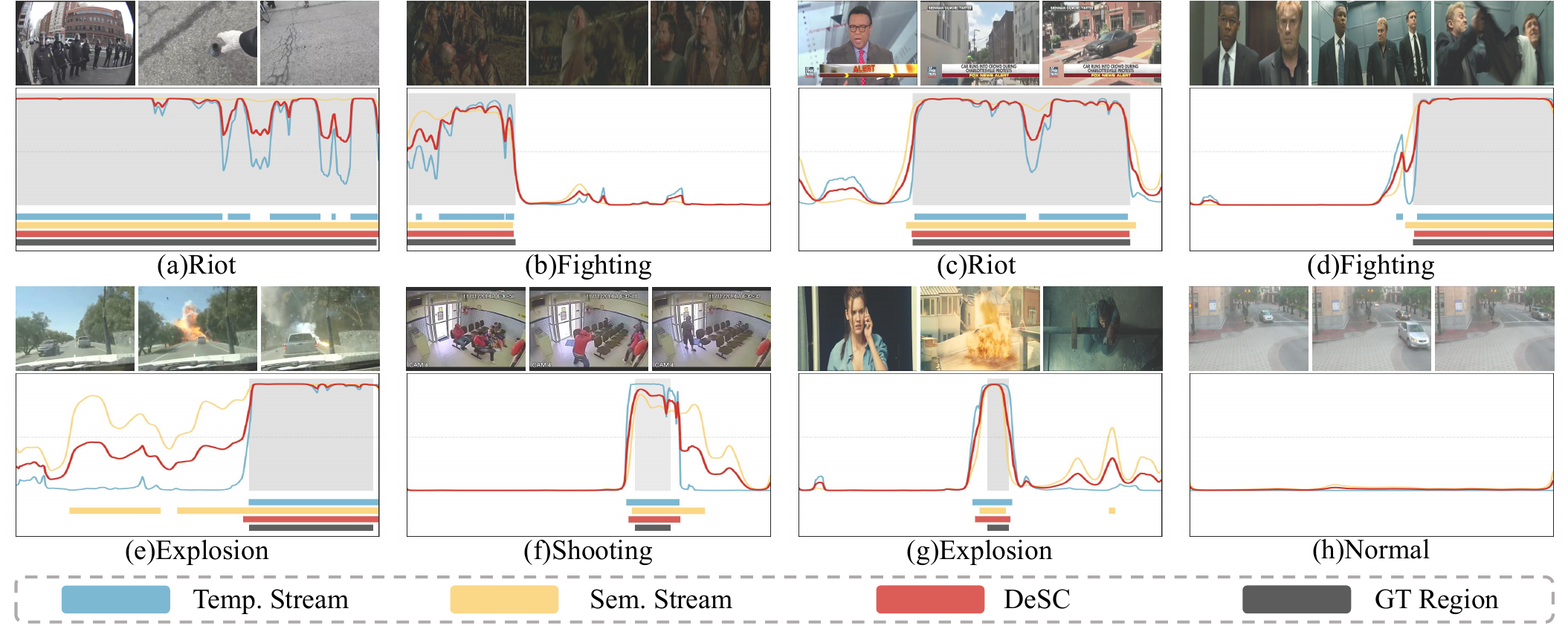}
  \caption{
Qualitative comparison on XD-Violence and UCF-Crime. 
Curves show anomaly scores from the Temporal Sensitivity Stream (\textcolor{c_temporal}{\textbf{blue}}), 
the Semantic Consistency Stream (\textcolor{c_semantic}{\textbf{yellow}}), 
and their fusion in DeSC (\textcolor{c_ours}{\textbf{red}}). 
Color bars below display binary detections aligned with the ground truth (\textcolor{c_gt}{\textbf{grey}}).
}
  \label{fig:qualitative}
\end{figure*}
\subsection{Implementation Details}


We use pre-trained CLIP ViT-B/16 to extract 512-D visual features. The temporal sensitivity stream comprises seven temporal convolutional and one graph transformer layers, optimized at a $1\times10^{-3}$ learning rate with a natural constant time-scale $\tau$. The semantic consistency stream consists of one transformer and three graph convolutional layers, trained with a robust $5\times10^{-5}$ learning rate and five Gaussian components ($K$). We set $\lambda$ as 0.7 for UCF-Crime and 0.1 for XD-Violence. Inference employs a 256-frame sliding window on a single NVIDIA RTX 4090 GPU.

\subsection{Comparison with State-of-the-art Methods}

We compare DeSC with state-of-the-art methods and categorize approaches into visual-only methods and vision–language methods as summarized in Table~\ref{tab:sota_comparison}. On the UCF-Crime dataset, visual-only models that rely on I3D features generally produce AUC scores below 87\%. Recent vision–language methods, including VadCLIP and STPrompt, incorporate semantic priors and have improved performance to about 88\%. However, these methods remain limited by the optimization conflict in unified frameworks. DeSC addresses this limitation and achieves a new state-of-the-art AUC of \textbf{89.37\%}. This improvement of \textbf{1.29\%} over the previous best method STPrompt demonstrates the effectiveness of decoupled sensitivity–consistency framework in surveillance environments.

On the XD-Violence dataset, which contains diverse videos from movies and online platforms, DeSC further shows strong robustness. Existing leading methods such as VadCLIP and CLIP-TSA achieve an AP of around 84\% through using CLIP features. DeSC attains an AP of \textbf{87.18\%}, outperforming VadCLIP by \textbf{2.67\%} and Holmes-VAD by \textbf{2.22\%}. This consistent advantage across both surveillance videos in UCF-Crime and heterogeneous internet videos in XD-Violence confirms that the decoupled optimization paradigm effectively leverages the potential of temporal modeling modules and provides a significant improvement over traditional joint training strategies.


\subsection{Ablation Studies}


\textbf{Necessity of Decoupled Optimization.}
This work is motivated by the observation that jointly optimizing sensitivity and stability within a single architecture causes gradient conflicts. To examine this hypothesis, we compare a unified architecture with the two decoupled streams in DeSC. The unified model merges the parallel temporal convolutional network and the Gaussian mixture prior module into one network and trains all components simultaneously. The results in Table~\ref{tab:ablation_optimization} show that this joint training strategy yields an AUC of 86.18\% on UCF-Crime, which is notably lower than the performance of either decoupled stream trained independently. The temporal sensitivity stream achieves 88.46\% AUC, and the semantic consistency stream achieves 88.35\% AUC. These results support the hypothesis that decoupling the optimization process enables each stream to achieve its best performance without being constrained by conflicting gradients.

\textbf{Contribution of Each Stream.}
Table~\ref{tab:ablation_optimization} further illustrates the complementary nature of the two streams. The temporal sensitivity stream performs well on detecting high-frequency transient anomalies but may respond to noise. The semantic consistency stream suppresses background noise through its Gaussian prior formulation yet may miss abrupt events. The collaborative inference mechanism combines their strengths and improves the performance to 89.37\% on UCF-Crime and 87.18\% on XD-Violence. This gain over the best individual stream confirms that the two streams learn distinct and complementary anomaly representations.

\textbf{Effectiveness of Collaborative Inference.}
We analyze the impact of the test-time augmentation strategy used in collaborative inference. Table~\ref{tab:ablation_inference} compares the basic ensemble method with the proposed sliding window approach. The basic ensemble averages the predictions from the two streams using standard non-overlapping video segments and produces an AUC of 89.13\%. Incorporating the sliding window strategy alleviates the boundary artifacts introduced by fixed-length segmentation and further raises the AUC to 89.37\%. This improvement demonstrates the benefit of smooth temporal transitions during inference.

\textbf{Effectiveness of Internal Designs.}
We examine the architectural decisions within each stream through component-level ablations, as summarized in Table~\ref{tab:ablation_internal}.
For the temporal sensitivity stream, we compare the proposed parallel TCN and GT architecture with variants that use only TCN or only GT. The results show that the single-module variants achieve 85.80\% and 86.26\% AUC on UCF-Crime, which are notably lower than the full parallel design. The combined architecture achieves 88.46\% AUC, demonstrating that capturing local high-frequency signals through TCN together with global dependencies through GT is beneficial for enhancing sensitivity to transient anomalies.
For the semantic consistency stream, we analyze the influence of multi-modal Gaussian Mixture Prior. Table~\ref{tab:ablation_internal} indicates that removing the GMP regularization reduces performance from 88.35\% to 87.70\% on UCF-Crime. This decline confirms that GMP effectively enforces temporal coherence and suppresses noise, acting as an learnable low-pass filter for modeling sustained anomalies.


\subsection{Qualitative Analyses}
To illustrate how DeSC balances sensitivity and stability in practice, Fig.~\ref{fig:qualitative} presents qualitative comparisons of frame-level anomaly scores. 
%
For sustained anomalies such as riot and fighting shown in the first row, the temporal sensitivity stream often produces fragmented responses when visual changes become subtle. In contrast, the semantic consistency stream, constrained by the Gaussian mixture prior, maintains a smooth and continuous prediction that reliably covers the entire abnormal event, enabling DeSC to fill the missing segments.
For transient events such as explosion and shooting in the second row, the semantic stream may exhibit over-smoothing or false positives in normal regions due to ambiguous semantics, while the temporal stream provides sharp and localized responses to abrupt changes. DeSC suppresses these semantic false alarms by leveraging low temporal confidence and achieves higher precision.
When both streams align, as in the fighting case Fig.~\ref{fig:qualitative}(d), the fused score increases significantly. In normal videos, both streams consistently remain low, confirming the robustness of DeSC against false positives.

\section{Limitaions And Future Work}\label{sec:limitations}
Although DeSC effectively resolves the sensitivity–stability conflict, its decoupled design increases training costs and model size compared to unified approaches, potentially hindering edge deployment. Future work will focus on distilling the dual-stream knowledge into a single lightweight model to reduce overhead and exploring advanced video–language backbones to enhance open-world adaptability.
\section{Conclusions}

This paper addresses the sensitivity–stability trade-off in WSVAD and identifies that the conflicting optimization demands for transient and sustained anomalies significantly limit the performance of unified frameworks.
To overcome this limitation, we introduce DeSC, a decoupled framework that trains two specialized streams in independent optimization spaces. The temporal sensitivity stream is designed to capture high-frequency transient anomalies, and the semantic consistency stream maintains long-term coherence. 
Their complementary strengths are integrated through a collaborative inference strategy that mitigates the limitations of each stream. 
Extensive experiments confirm that this decoupled paradigm achieves new state-of-the-art performance on UCF-Crime and XD-Violence, demonstrating the effectiveness of decoupling and fusing sensitivity and stability for robust anomaly detection.

\bibliographystyle{IEEEbib}
\bibliography{icme2026references}


\end{document}